%% file: main.tex
\documentclass{article}

\PassOptionsToPackage{numbers,compress}{natbib}

\usepackage[preprint]{neurips_2026}

\input{preamble.tex}
\input{macros.tex}

\title{\our{}: Evolutionary Reward Function Design for Robot Navigation with Large Language Models}

\author{%
  Zhikai Zhao$^{1}$\thanks{Equal contribution.}\ \ ,\ Chuanbo Hua$^{1}$\footnotemark[1]\ \ ,\ Federico Berto$^{2}$,\ Zihan Ma$^{1}$,\\
  \textbf{Kanghoon Lee}$^{1}$,\ \textbf{Jiachen Li}$^{3}$,\ \textbf{Jinkyoo Park}$^{1,4}$ \\[2pt]
  \normalfont\small $^{1}$KAIST \quad $^{2}$Radical Numerics \quad $^{3}$UC Riverside \quad $^{4}$Omelet \quad AI4CO\thanks{Works from the AI4CO open research community.}
}

\begin{document}

\renewcommand{\thefootnote}{\fnsymbol{footnote}}
\maketitle
\renewcommand{\thefootnote}{\arabic{footnote}}
\setcounter{footnote}{0}

\vspace{-3mm}

\input{sections/00-abstract.tex}
\input{sections/01-introduction.tex}
\input{sections/02-related-works.tex}
\input{sections/03-preliminaries.tex}
\input{sections/04-methodology.tex}

\input{sections/05-experiment.tex}
\input{sections/06-conclusion.tex}

\bibliographystyle{plainnat}
\bibliography{bib/refs}

\newpage
\appendix

\input{appendix/a-algorithm.tex}
\input{appendix/b-hyperparameters.tex}
\input{appendix/c-prompts.tex}

\end{document}

%% file: preamble.tex
\usepackage[utf8]{inputenc}
\usepackage[T1]{fontenc}
\usepackage{upquote}

\usepackage{microtype}
\usepackage{graphicx}
\usepackage{subcaption}
\usepackage{wrapfig}
\usepackage{xcolor}

\usepackage{booktabs}
\usepackage{multirow}
\usepackage{diagbox}
\usepackage{nicefrac}

\usepackage{url}
\usepackage{hyperref}

\usepackage{amsmath}
\usepackage{amssymb}
\usepackage{amsfonts}
\usepackage{mathtools}
\usepackage{amsthm}

\usepackage{algorithm}
\usepackage{algorithmic}

\usepackage[textsize=tiny]{todonotes}

\usepackage{fancyvrb}
\usepackage{fvextra}
\fvset{breaklines=true, breakanywhere=true}

\usepackage{tcolorbox}
\tcbuselibrary{listings,skins,breakable}

\usepackage[capitalize,noabbrev]{cleveref}

\theoremstyle{plain}

\theoremstyle{definition}

\theoremstyle{remark}

\definecolor{PromptBG}{RGB}{248,248,248}
\definecolor{PromptFrame}{RGB}{200,200,200}
\definecolor{PromptTitle}{RGB}{40,40,40}
\definecolor{PromptAccent}{RGB}{23,90,170}
\definecolor{PromptAccent2}{RGB}{160,40,40}

\lstdefinestyle{promptstyle}{
  backgroundcolor=\color{PromptBG},
  basicstyle=\small\ttfamily\color{black},
  frame=single,
  rulecolor=\color{PromptFrame},
  framesep=4pt,
  xleftmargin=6pt, xrightmargin=6pt,
  aboveskip=0.9\baselineskip, belowskip=0.9\baselineskip,
  showstringspaces=false, showspaces=false, showtabs=false,
  numbers=none, numbersep=6pt,
  breaklines=true, breakatwhitespace=false,
  columns=flexible, upquote=true, tabsize=4,
  moredelim=**[is][\bfseries\color{PromptAccent}]{§}{§},
  inputencoding=utf8, extendedchars=true
}

\lstnewenvironment{prompt}[1][]
  {\lstset{style=promptstyle, caption={#1}}}
  {}

\newtcolorbox{promptbox}[2][]{%
  enhanced,
  breakable,
  colback=PromptBG,
  colframe=PromptFrame,
  coltitle=PromptTitle,
  fonttitle=\bfseries\small,
  title={#2},
  attach boxed title to top left={yshift=-2mm, xshift=2mm},
  boxed title style={colback=gray!10, colframe=PromptFrame, sharp corners},
  left=2mm, right=2mm, top=2mm, bottom=2mm,
  arc=2mm,
  listing only,
  listing options={style=promptstyle},
  #1
}

\crefname{lstlisting}{listing}{listings}
\Crefname{lstlisting}{Listing}{Listings}

\lstdefinestyle{heuristicstyle}{
    backgroundcolor=\color{gray!5},
    commentstyle=\color{green!60!black},
    keywordstyle=\color{blue!70!black}\bfseries,
    numberstyle=\tiny\color{gray!70},
    stringstyle=\color{purple!70!black},
    basicstyle=\small\ttfamily\color{black},
    breakatwhitespace=false,
    breaklines=true,
    captionpos=b,
    keepspaces=true,
    showspaces=false,
    showstringspaces=false,
    showtabs=false,
    tabsize=4,
    frame=single,
    rulecolor=\color{gray!50},
    framesep=3pt,
    frameround=tttt,
    numbersep=6pt,
    xleftmargin=10pt,
    xrightmargin=10pt,
    aboveskip=1.0\baselineskip,
    belowskip=1.0\baselineskip,
    upquote=true,
    columns=flexible,
    keepspaces=true,
    mathescape=true,
    escapeinside={(*@}{@*)},
    morecomment=[l]\#,
    morekeywords={import, from, as, def, class, return, yield, for, while, if, elif, else, try, except, finally, with, lambda, pass, break, continue, and, or, not, is, in, raise, assert},
    emph={self, None, True, False, np, pd, plt, torch, tf, sklearn},
    emphstyle=\color{orange!80!black}\bfseries,
    literate=
        {-}{-}1
        {=>}{$\Rightarrow$ }3
        {->}{$\rightarrow$ }3
        {...}{$\ldots$ }3,
    inputencoding=utf8,
    extendedchars=true,
    lineskip=-0.1pt,
    fontadjust=true
}

%% file: macros.tex
\newcommand{\our}{EvoNav}

\newcommand{\std}[1]{\text{\ensuremath{\pm}\,#1\%}}
\newcommand{\stdnp}[1]{\text{\ensuremath{\pm}\,#1}}

%% file: sections/00-abstract.tex
\begin{abstract}
    Robot navigation is a crucial task with applications to social robots in dynamic human environments. While Reinforcement Learning (RL) has shown great promise for this problem, the policy quality is highly sensitive to the specification of reward functions. Hand-crafted rewards require substantial domain expertise and embed inductive biases that are difficult to audit or adapt, limiting their effectiveness and leading to suboptimal performance.
    In this paper, we propose \our{}, an evolutionary framework that automates the design of robot navigation reward functions via large language models (LLMs). To overcome prohibitively costly policy training, \our{} evaluates each candidate proposal from the LLM via a progressive three-stage warm-up-boost procedure.
    \our{} advances from analytical proxies with low-cost surrogates, such as small datasets and analytic rules, to lightweight rollouts and, finally, to full policy training, enabling computationally efficient exploration under effective feedback. Experiment results show that \our{} produces more effective navigation policies than manually designed RL rewards and state-of-the-art reward design methods.
\end{abstract}
\vspace{-3mm}

%% file: sections/01-introduction.tex
\section{Introduction}
\label{sec:introduction}

\begin{wrapfigure}{r}{0.5\textwidth}
    \vspace{-8mm}
    \centering
    \includegraphics[width=0.48\textwidth]{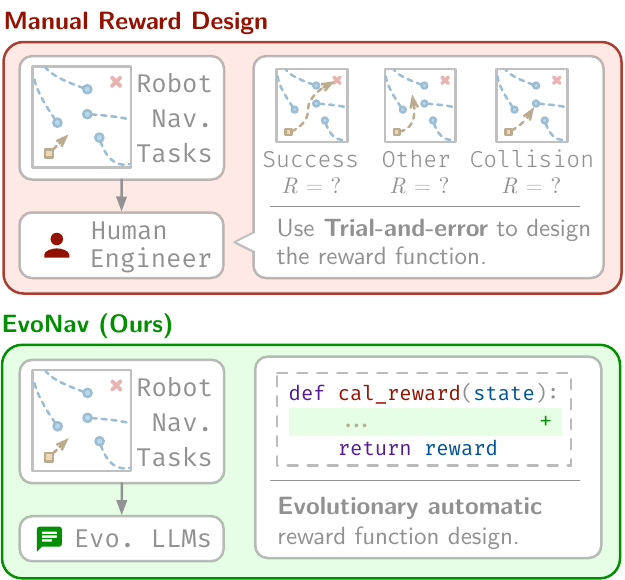}
    \caption{Motivation for \our{}. Traditional manual reward function design (top) relies on human experts and extensive trial-and-error. \our{} (bottom) automates reward function design through an evolutionary framework guided by LLMs.}
    \label{fig:motivation}
    \vspace{-8mm}
\end{wrapfigure}


Robot navigation among dynamic agents is central to service robotics and autonomous driving, yet remains challenging due to implicit interactions, partial observability, and error propagation \citep{mavrogiannis2021corechallengessocialrobot,neupane2024securityconsiderationsairoboticssurvey,le2024comprehensive,singamaneni2024survey}. While research has explored rule-based, optimization-based, and hybrid approaches \citep{matsumoto2024crowd, zhu2021rule, huajian2024sample, raj2024rethinking}, reinforcement learning (RL) is particularly suitable as learned policies produce real-time actions encoding complex interaction patterns \citep{jing2024two, zhou2025her, montero2025memory}.


Common practice relies on manually specified components (progress, collision penalties, social compliance) combined as weighted sums tuned through trial-and-error \citep{ng1999policy,dewey2014reinforcement}. This approach demands domain knowledge, contains implicit biases that are difficult to detect \citep{hadfield2016cooperative,skalse2022defining}, and does not scale across environments where priorities shift or constraints tighten.


Additionally, reward evaluation requires full policy training and evaluation, creating a slow, expensive feedback loop \citep{ma2023eureka,henderson2018deep,dulac2021challenges}. In settings like crowd navigation, where simulation is already costly and safety metrics require meaningful experience, exhaustive search over reward designs is impractical.

As shown in \cref{fig:motivation}, large language models (LLMs) offer a new opportunity to address this challenge: by encoding broad world knowledge, LLMs can decompose natural language descriptions into structured specifications, and support iterative improvement through critique and revision. These capabilities align with the needs of reward design, where the space of plausible components is large and interpretability matters \citep{ma2023eureka, sun2025large,guo2024connecting,liu2024evolution}. However, directly attaching a language model-driven evolutionary process to reinforcement learning introduces a fundamental obstacle. In typical LLM-based algorithm design tasks, the evaluation of generated code is relatively inexpensive, as the produced heuristics can be directly executed and assessed without significant computational overhead \citep{liu2024llm4ad}. In the reward function design task, each candidate reward must be evaluated through policy learning to assess its downstream effects on success, safety, and efficiency. The evaluation budget is therefore the bottleneck for iterative evolution. Without an efficient strategy that rationalizes computation, naive evolutionary exploration by training a policy for every candidate is infeasible.

We introduce \our{}, an evolutionary framework that automates reward design for robot navigation through a progressive three-stage framework. We first rapidly explore the reward function space starting from a seed function via evolution guided by analytical rules computed on pre-collected trajectory datasets as a warm-up stage. Next, we transition to the boost phase, where evolved candidates undergo iterative refinement through lightweight proxy model training, with the LLM making targeted adjustments based on multi-objective behavioral metrics. Finally, the best candidates are refined through limited full-scale training rounds to achieve optimal performance. This warm-up-boost pipeline efficiently allocates expensive computation to promising candidates while maintaining systematic exploration through low-cost proxy evaluations.

The main contributions of this paper are as follows. (1) We propose \our{}, a language model-driven framework for automated reward design in robot navigation. (2) We introduce a progressive three-stage framework that reduces the number of full trainings required during search. (3) We validate our approach on dense crowd navigation benchmarks, where it improves the success rate and reduces collisions/timeouts over strong navigation references and outperforms state-of-the-art reward-design baselines.

%% file: sections/02-related-works.tex
\section{Related Works}
\label{sec:related-works}

\paragraph{Robot Navigation Methods}
Robot navigation in dynamic environments remains a fundamental challenge in robotics. Traditional approaches rely primarily on heuristic-based methods, including the Social Force Model and velocity-based techniques such as Velocity Obstacles, Reciprocal Velocity Obstacles, and Optimal Reciprocal Collision Avoidance \citep{Helbing_1995,van2008reciprocal,van2011reciprocal}. These methods provide interpretability and computational efficiency without requiring training, but depend on manually crafted rules and parameters that often fail to generalize across diverse scenarios \citep{singamaneni2024survey,mavrogiannis2021corechallengessocialrobot}.


Recent advances in deep reinforcement learning have shown significant promise, with methods like Socially Attentive Reinforcement Learning demonstrating superior performance by explicitly modeling human-robot interactions through attention mechanisms \citep{chen2019crowd,everett2018motion}. Graph-based approaches such as Decentralized Structural-RNN further improve navigation by capturing spatial-temporal relationships and incorporating human intention prediction \citep{liu2021decentralized,liu2023intentionawarerobotcrowd}. Complementary work on relational reasoning over group structure for trajectory prediction and social navigation \citep{li2022evolvehypergraph,li2024multiagent} and on heuristic-informed policies that scale to large numbers of agents \citep{tang2024himap} further underscores the importance of structured inductive biases in crowded, interactive environments. Despite these advances, the navigation policy quality remains highly sensitive to reward function design, motivating our work on automated reward synthesis.

\vspace{-2mm}

\paragraph{Reward Function Design}
Recent work has begun to leverage large language models for automated reward design in reinforcement learning. Early approaches use LLMs to provide direct reward signals \citep{kwon2023reward} or translate natural language into rewards through predefined APIs \citep{yu2023language}, but these limit interpretability or expressiveness. More recent frameworks such as Text2Reward \citep{xie2023text2reward}, Eureka \citep{ma2023eureka}, and Auto MC-Reward \citep{li2024auto} represent rewards as executable code, but require full policy training to evaluate each candidate, making them computationally expensive. CARD \citep{sun2025large} introduces Trajectory Preference Evaluation to filter candidates before training, yet still necessitates full training for each passing candidate.

A critical limitation shared by these methods is their evaluation strategy: most treat reward assessment as generating a candidate, training a complete policy, and measuring performance. This becomes prohibitive when policy training is expensive, as in robot navigation where realistic simulations require substantial compute \citep{henderson2018deep,dulac2021challenges}. Unlike heuristic algorithm design, where candidates can be evaluated by direct execution, reward functions reveal their quality only through downstream policy behavior, creating a computational bottleneck that existing work does not systematically address. A related line of work uses LLM-driven evolution to synthesise executable heuristics in other domains where evaluation is cheap, including combinatorial optimisation \citep{ye2024reevo,hottung2025vrpagent}, trajectory prediction \citep{zhao2025trajevo}, and energy forecasting \citep{lin2025buildevo}; our work imports this paradigm into reward design, where the expensive-evaluation bottleneck is precisely what makes a progressive-fidelity search necessary.

%% file: sections/03-preliminaries.tex
\section{Preliminaries}
\label{sec:preliminary}

\begin{figure*}[t!]
    \centering
    \includegraphics[width=\linewidth]{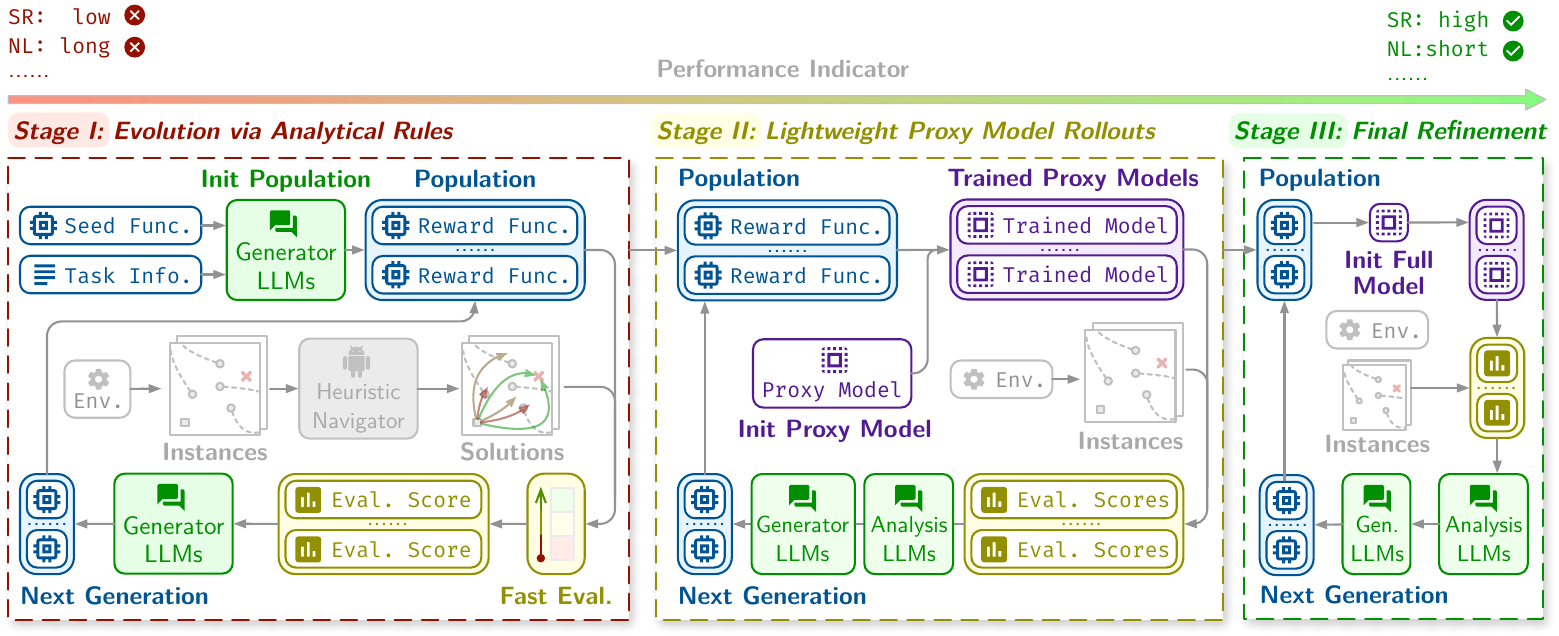}
    \caption{Overview of \our{}’s three-stage pipeline: \textbf{(Left)} Stage I evolves reward candidates from a seed via analytical rule-based screening on pre-collected trajectories, requiring no policy training; \textbf{(Middle)} Stage II refines candidates with lightweight proxy-policy training and LLM-guided edits using multi-objective metrics; \textbf{(Right)} Stage III performs limited full-policy training and evaluation to select the final reward, concentrating compute on the most promising candidates.}
    \label{fig:overview}
    \vspace{-4mm}
\end{figure*}

\paragraph{Problem Definition}
We consider planar crowd navigation in discrete time $t\in\{0,\ldots,T-1\}$ within a workspace $\mathcal{W}\subset\mathbb{R}^2$. A single robot with disc footprint radius $r_0$ moves among $H$ humans indexed by $h\in\{1,\ldots,H\}$, where the ground truth position of human $h$ at time $t$ is $p_h(t)\in\mathbb{R}^2$ and the robot position is $x_t\in\mathbb{R}^2$. The robot starts from $x_0$ and aims to reach a goal $g\in\mathcal{W}$ while avoiding collisions and unnecessary discomfort. Control is holonomic with action $a_t=(v_x(t),v_y(t))\in\mathcal{A}=\{a\in\mathbb{R}^2:\|a\|\le v_{\max}\}$ and first order kinematics $x_{t+1}=x_t+\Delta t\, a_t$. A collision at time $t$ occurs when $\min_{h}\|x_t-p_h(t)\|\le r_0+r_h$, where $r_h$ is the footprint radius of human $h$. Success is declared when $\|x_t-g\|\le \varepsilon_{\text{goal}}$. Episodes terminate on success, collision, or at horizon $T$. At each time $t$, the agent receives a state $s_t=[e_t, h_t, m_t] \in \mathcal{S}$, where $e_t$ contains ego quantities (e.g., robot position $x_t$ and feasible action constraints), $h_t$ includes measurements of humans (e.g., positions $\{p_h(t)\}_{h=1}^H$ for $H$ humans), and $m_t$ denotes auxiliary model outputs such as predicted human trajectories $\{\hat{p}_{h,k}(t)\in\mathbb{R}^2\}_{h=1..H,k=1..K}$ for $K$ future steps, with $\mathcal{S}$ being the state space. The objective is to navigate the robot to $g$ safely and efficiently under these interactive dynamics.

We cast this navigation task as a finite-horizon Markov Decision Process (MDP) $\mathcal{M}=(\mathcal{S},\mathcal{A},P,R_{\theta},\gamma,\rho_{0})$ whose per-step reward $R_{\theta}$ is parameterised by $\theta$ and optimised by \our{}; the full MDP formulation and the environment parameter set are deferred to \cref{sec:mdp} in the appendix.

%% file: sections/04-methodology.tex
\section{Methodology}
\label{sec:methodology}
We present \our{}, an evolutionary framework that leverages LLMs to automatically design reward functions for robot navigation while managing the high cost of policy evaluation. The key insight is to organize the search process into three progressive stages of increasing fidelity, where inexpensive proxies identify poor candidates early, allowing the full training budget to concentrate on promising reward specifications.

\vspace{-2mm}
\subsection{Overview of EvoNav}



\cref{fig:overview} illustrates the overview of \our{} framework. The design is motivated by the observation that reward quality can be assessed at multiple fidelity levels, from analytical signals to full training, enabling efficient search without sacrificing quality. Early stages rapidly filter candidates while later stages refine promising designs, allowing broader exploration than uniform full-training evaluation. The framework operates in three progressive stages with distinct objectives:

\paragraph{Stage I – Evolution via Analytical Rules (\cref{subsec:stage-1})}
Starting from a seed function, the LLM generates an initial population of $N$ candidate reward functions. Over $G_1$ generations, candidates evolve through mutation, crossover, and reflection operations, guided by fast rank-correlation scores computed on pre-collected trajectory datasets. This stage explores the reward function space efficiently without policy training as a warm-up.

\paragraph{Stage II – Lightweight Proxy Model Rollouts (\cref{subsec:stage-2})}
The evolved reward function candidates from Stage~I undergo iterative refinement over $G_2$ generations. Each round trains lightweight proxy policies and evaluates multi-objective metrics. We utilize LLM to analyze these metrics and make adjustments to the design of each reward function. This stage exploits local structure through intensive optimization at a relevant low computational cost.

\paragraph{Stage III – Final Refinement (\cref{subsec:stage-3})}
Finally, the reward candidates refined in Stage~II enter Stage~III for full-policy training across $G_3$ refinement rounds. Each round trains a complete policy and evaluates comprehensive performance metrics to align the reward with end-to-end behavior. Given the substantial computational cost of full-policy training, we restrict $G_3$ to a small value in practice, for computational and budgetary reasons.

This progressive organization implements an explore–exploit strategy: Stage~I explores diverse reward structures via evolution using inexpensive signals; Stage~II exploits promising candidates through iterative local optimization at low cost; and Stage~III performs final alignment with high-fidelity, accurate signals. The LLM receives detailed performance metrics at each stage to guide both evolution (Stage~I) and refinement (Stages~II and~III).

\subsection{LLM-Based Generation Component}
We present the LLM-driven generator that proposes, evolves, and selects reward functions throughout \our{}’s three-stage pipeline.

\paragraph{Reward Representation}
Each candidate reward function $r_\theta$ is represented as executable Python code. This code-based representation promotes interpretability by allowing explicit component definitions while enabling the LLM to explore diverse mathematical formulations and weighting schemes.
The generated code adheres to a standardized interface template, ensuring compatibility with the RL training procedure. Specifically, each function has the signature \texttt{cal\_reward($s_t$)} used consistently across all stages of evaluation. The function returns a scalar reward value for the current frame.

\paragraph{Initial Population}
We initialize the search by seeding the generator LLM with a compact task specification (state/action interface, metrics, and constraints) and a simplified, general seed reward derived from CrowdNav++ \citep{liu2023intentionawarerobotcrowd}. The seed follows a value-weighted template (e.g., a large terminal bonus for success, a terminal penalty for collision, and a shaping term such as negative distance-to-goal), chosen for clarity and generality rather than optimality. The LLM then produces an initial population of $N$ diverse candidates by perturbing this template and proposing a small number of variants, which serves as the starting point for Stage~I search.

\paragraph{Evolutionary Operations}
The evolutionary loop follows a generate, evaluate, and then select cycle with lightweight reflective summaries to guide subsequent proposals, in line with recent reflective-evolution paradigms \citep{ye2024reevo,zhao2025trajevo,hottung2025vrpagent}. In \emph{mutation}, given a parent reward and its performance profile, the LLM proposes modifications such as introducing new insights and components, adjusting component weights or schedules, or altering functional forms. In \emph{crossover}, the LLM synthesizes a new candidate from two high-performing parents by combining complementary strengths (for example, merging a strong collision-avoidance term from one with an efficient progress-shaping term from the other). To preserve diversity and avoid premature convergence, we also introduce occasional random restarts. Brief reflective notes, accumulated across generations, provide concise textual guidance that biases future mutations and crossovers without adding heavy optimization overhead.

\paragraph{Structured Feedback}
The LLM receives different feedback across stages to guide evolution and refinement. In Stage~I, the LLM receives correlation scores for all candidates after each generation, which guides evolutionary operations: mutation addresses individual weaknesses, crossover combines complementary strengths, and random initialization maintains diversity. In Stages~II and III, after each training round, the LLM receives multi-objective metrics for each candidate and analyzes performance profiles to make targeted code adjustments addressing specific weaknesses. The LLM uses raw metric values rather than rankings to guide modifications, enabling dimension-specific improvements.

\subsection{Progressive-Fidelity Reward Search}
We frame reward design as a search problem under expensive evaluation. EvoNav implements a progressive-fidelity reward search strategy, in which inexpensive proxies are used to triage candidates early, and increasingly faithful evaluations are applied only to a small subset. This design prioritizes rank preservation over absolute accuracy and enables efficient allocation of training compute.

EvoNav is a concrete instantiation of this progressive-fidelity reward search strategy for robot navigation. We detail the three stages of the \our{} framework, which constitute a progressive evaluation pipeline. The complete pseudocode is provided in \cref{alg:evonav} in the appendix.

\subsubsection{Stage I: Evolution via Analytical Rules}
\label{subsec:stage-1}

\begin{wrapfigure}{r}{0.55\textwidth}
    \vspace{-\intextsep}
    \centering
    \includegraphics[width=0.52\textwidth]{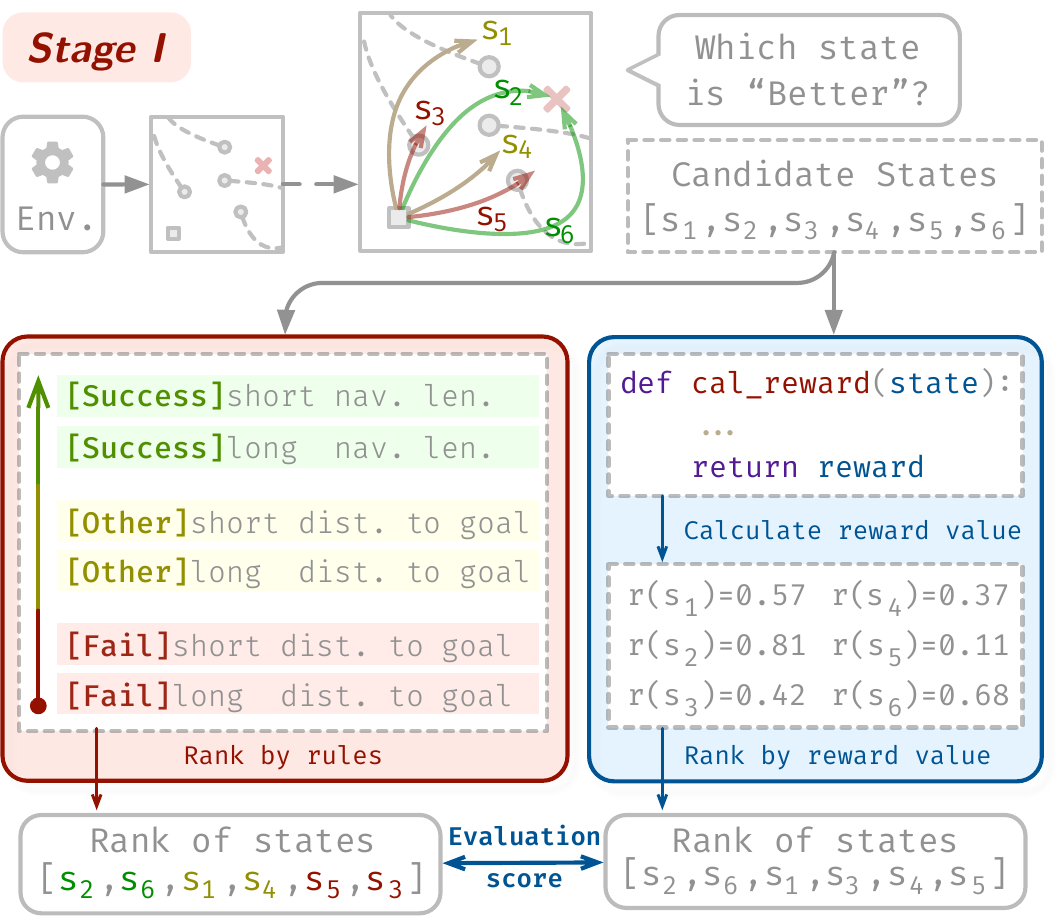}
    \caption{Illustration of the analytical rules in Stage I}
    \label{fig:stage1}
    \vspace{-\intextsep}
\end{wrapfigure}

\cref{fig:stage1} illustrates the Stage~I evaluation details. Stage~I rapidly explores the reward function space through evolution guided by analytical rules. The LLM first generates $N$ initial candidates as the initial population $\mathcal{P}^{(0)}$ from a seed function using structured prompting. At each generation $g$, we evaluate all candidates using rank correlation on a pre-collected dataset $\mathcal{D}$ containing $M$ navigation scenarios, each with $N_{\text{traj}}$ diverse trajectories exhibiting different navigation behaviors. For each frame $f$ in scenario $j$, we establish two orderings over trajectories: a rules-based ranking $rank_{\text{rules}}^{(j,f)}$ that prioritizes success over progress over collision, and a reward-based ranking $rank_{r_\theta}^{(j,f)}$ ordered by cumulative reward from each candidate function. Note that these rules are derived directly from the target measurement matrix, eliminating the need for a domain expert to design them manually. The correlation score aggregates alignment across all frames and scenarios:
\begin{equation}
\text{Score}_1(r_\theta) = \frac{1}{M} \sum_{j=1}^{M} \frac{1}{F_j} \sum_{f=1}^{F_j} \text{Corr}(rank_{\text{rules}}^{(j,f)}, rank_{r_\theta}^{(j,f)})
\end{equation}
where $F_j$ denotes the number of frames in scenario $j$. $\text{Corr}(\cdot, \cdot)$ is the Spearman rank correlation coefficient. A higher $Score_1$ indicates that the reward function's preferences align with quality rules.

After computing scores for all candidates, we produce a ranking $\mathcal{R}_1^{(g)}$ by sorting. The LLM then uses these scores as feedback to generate the next generation $\mathcal{P}_{next}$ through three operations: mutation, crossover, and random initialization. This evolutionary process repeats for $G_1$ generations, producing $N$ well-evolved candidates $\mathcal{P}_{\text{evolved}}$ that proceed to Stage~II. The computational cost is negligible, requiring only minutes of CPU time per generation.

\subsubsection{Stage II: Lightweight Proxy Model Rollouts}
\label{subsec:stage-2}

\input{tables/main}

Stage~II transitions from exploration to exploitation through iterative refinement. The $N$ evolved candidates from Stage~I undergo $G_2$ rounds of training and adjustment. In each round, we initialize a fresh lightweight policy $\pi_\phi^{\text{proxy}}$ for each candidate $r$ and train from scratch for $K_2$ gradient steps (\cref{tab:stage2_params}). We then collect episodes with horizon steps and measure multiple objectives $M(r)$.

After evaluating all candidates, the LLM analyzes their performance profiles and makes targeted code adjustments to address identified weaknesses. This iterative refinement continues for $G_2$ rounds, enabling local optimization guided by LLM reasoning. After the final round, the LLM produces a ranking through a comprehensive multi-objective evaluation of all metrics.

\subsubsection{Stage III: Final Refinement}
\label{subsec:stage-3}

Stage~III performs final polishing through full-scale policy training. The $N$ refined candidates from Stage~II undergo $G_3$ rounds of training and adjustment. In each round $k$, we initialize a fresh full policy $\pi_\phi^{\text{full}}$ for each candidate $r$ and train from scratch for $K_3$ environment steps (\cref{tab:stage3_params}). We then evaluate each policy on test episodes and measure the same metrics $M(r)$ as Stage~II.

After evaluating all candidates, the LLM analyzes their comprehensive performance profiles and makes final targeted adjustments to each candidate's code. This iterative refinement continues for $G_3$ rounds. After the final round, the LLM produces the final ranking $\mathcal{R}_3$ through multi-objective evaluation. This ranking provides the most reliable quality ordering, as it is based on full-scale training with comprehensive performance signals.

\subsubsection{Proxy consistency.}
\our{} relies on a practical assumption: reward functions that rank highly under inexpensive proxies (analytic scores and proxy-policy rollouts) tend to remain strong when assessed with the full model. This is analogous to scaling-law intuition \citep{kaplan2020scaling,hoffmann2022training}, where smaller models and shorter horizons already capture a signal that preserves useful ordering information for larger, more expensive evaluations \citep{biagiola2023testing,chen2024cost}. Rather than stating a formal hypothesis, we assess this transfer empirically by computing Spearman rank correlations between stage-wise rankings (Stage~I vs.~II, Stage~II vs.~III) and by reporting top-$k$ preservation rates. Consistently positive correlations and high preservation indicate that early stages triage the search effectively, concentrating full-training budget on promising candidates \citep{peherstorfer2018survey,forrester2008engineering}.

%% file: tables/main.tex
\begin{table}[ht]
\centering
\caption{Performance comparison of navigation methods across multiple metrics. Results are reported for scenarios with and without human randomization. \our{} consistently outperforms state-of-the-art reward design methods (Eureka, CARD) and learning-based baselines (CrowdNav++), achieving the highest success rates and lowest collision/timeout rates.}
\vspace{2mm}
\label{tab:main}
\small
\setlength{\tabcolsep}{4pt}
\makebox[\textwidth][c]{%
\begin{tabular}{l|l|ccc|cccc}
\toprule
& & \multicolumn{3}{c|}{\textbf{Main}} & \multicolumn{4}{c}{\textbf{Secondary}} \\
 \cmidrule(lr){3-5} \cmidrule(lr){6-9}
& \diagbox{\textbf{Method}}{\textbf{Metrics}} & \textbf{SR}$\uparrow$ & \textbf{CR}$\downarrow$ & \textbf{TR}$\downarrow$ & \textbf{NT}$\downarrow$ & \textbf{PL}$\downarrow$ & \textbf{ITR}$\downarrow$ & \textbf{SD}$\uparrow$ \\
\midrule
\multirow{7}{*}{\rotatebox{90}{w/ random}}
& SF            & 29.30\% & 21.42\% & 49.55\% & 17.68 & 15.93 & 17.68 & 0.37 \\
& ORCA          & 69.11\% & 24.21\% &  6.68\%  & 19.61 & 17.67 & 19.61 & 0.38 \\
& DS\text{-}RNN & 64.14\% & 25.41\% & 10.45\%  & 23.91 & 19.63 & 23.91 & 0.34 \\
& CrowdNav++    & $89.35\std{0.68}$ & $7.88\std{0.54}$ & $2.77\std{0.13}$ & $15.03\stdnp{0.32}$ & $21.31\stdnp{0.48}$ & $4.18\stdnp{0.26}$ & $0.44\stdnp{0.01}$ \\
\cmidrule(lr){2-9}
& Eureka        & $86.71\std{0.74}$ & $10.03\std{0.60}$ & $3.26\std{0.15}$ & $15.63\stdnp{0.28}$ & $20.87\stdnp{0.37}$ & $7.26\stdnp{0.34}$ & $0.44\stdnp{0.01}$ \\
& CARD          & $91.35\std{0.69}$ & $6.12\std{0.48}$  & $2.53\std{0.12}$ & $15.15\stdnp{0.30}$ & $20.83\stdnp{0.41}$ & $7.20\stdnp{0.33}$ & $0.42\stdnp{0.01}$ \\
\cmidrule(lr){2-9}
& EvoNav        & $\mathbf{93.22}\std{0.61}$ & $\mathbf{4.47}\std{0.42}$ & $\mathbf{2.31}\std{0.10}$ & $15.26\stdnp{0.27}$ & $20.38\stdnp{0.36}$ & $7.71\stdnp{0.30}$ & $0.42\stdnp{0.01}$ \\
\midrule
\multirow{7}{*}{\rotatebox{90}{w/o random}}
& SF            & 34.11\% & 29.84\% & 36.05\% & 19.95 & 17.75 & 21.35 & 0.35 \\
& ORCA          & 78.10\% & 17.13\% &  4.77\%  & 15.87 & 18.53 & 26.04 & 0.36 \\
& DS\text{-}RNN & 67.88\% & 27.06\% &  5.06\%  & 20.06 & 25.42 & 13.31 & 0.37 \\
& CrowdNav++    & $94.03\std{0.66}$ & $4.88\std{0.45}$  & $1.09\std{0.08}$  & $17.64\stdnp{0.35}$ & $22.51\stdnp{0.40}$ & $3.06\stdnp{0.24}$ & $0.43\stdnp{0.01}$ \\
\cmidrule(lr){2-9}
& Eureka        & $92.71\std{0.62}$ & $5.89\std{0.46}$ & $1.40\std{0.10}$ & $17.49\stdnp{0.32}$ & $22.18\stdnp{0.38}$ & $7.71\stdnp{0.29}$ & $0.41\stdnp{0.01}$ \\
& CARD          & $94.33\std{0.58}$ & $4.71\std{0.40}$ & $0.96\std{0.08}$ & $17.52\stdnp{0.30}$ & $22.03\stdnp{0.36}$ & $7.47\stdnp{0.27}$ & $0.41\stdnp{0.01}$ \\
\cmidrule(lr){2-9}
& EvoNav        & $\mathbf{95.61}\std{0.55}$ & $\mathbf{4.01}\std{0.36}$ & $\mathbf{0.38}\std{0.06}$ & $17.01\stdnp{0.28}$ & $19.98\stdnp{0.33}$ & $8.89\stdnp{0.25}$ & $0.40\stdnp{0.01}$ \\
\bottomrule
\end{tabular}}
\end{table}

%% file: sections/05-experiment.tex
\section{Experiments}
\label{sec:experiment}

\subsection{Experiments Setup}
\label{sec:experiment-setup}

\paragraph{Environment}
We use a 2D continuous simulator on a square workspace $\mathcal{W}=[0,12]\times[0,12]\ \text{m}^2$ where a holonomic robot navigates among up to $H=20$ humans. The robot has radius $r_0$, sensing range $R_{\text{sense}}=5\ \text{m}$, and maximum speed $v_{\max}=1\ \text{m/s}$. Humans are controlled by an ORCA-based \citep{10.1007/978-3-642-19457-3_1} controller that reacts to other humans but not to the robot. We report results under two settings: \textit{with randomization} (humans may switch goals mid-episode, traits randomized with $v_{h,\max}\in[0.5,1.5]\ \text{m/s}$, $r_h\in[0.3,0.5]\ \text{m}$) and \textit{without randomization} (fixed traits, no mid-episode goal switches). Both yield dense, interactive scenes.

\paragraph{Baselines}
We compare against: \textit{1) Reference navigation methods}: SF \citep{Helbing_1995} and ORCA \citep{10.1007/978-3-642-19457-3_1} (classical reactive), DS-RNN \citep{liu2021decentralized} (graph-based learning), and CrowdNav++ \citep{liu2023intentionawarerobotcrowd} (attention-based RL). \textit{2) Automated reward design baselines}: Eureka \citep{ma2023eureka} and CARD \citep{sun2025large}. All methods use identical policy architectures, training protocols, and evaluation metrics to ensure fair comparison.

\paragraph{Configurations}
We implement \our{} with a population of $N=8$ candidates evolved over $G_1=10$ generations in Stage~I, evaluated on $M=100$ pre-collected scenarios with $N_{\text{traj}}=10$ trajectories each. In Stage~II, we use a lightweight proxy A2C trained for $K_2=8{,}000$ steps across $G_2=16$ refinement rounds. In Stage~III, we train full policies for $K_3=10^7$ steps over $G_3=3$ rounds. All methods use identical CrowdNav policy architectures \citep{liu2023intentionawarerobotcrowd} with PPO optimization, implemented in PyTorch on NVIDIA A6000 GPUs. We use the open-source gpt-oss-120B \citep{agarwal2025gpt} as the LLM served locally via vLLM \citep{kwon2023efficient}. Critically, \our{}, Eureka, and CARD in \cref{tab:main} differ only in their reward functions—all use the same CrowdNav architecture and training protocol, ensuring performance differences reflect reward design quality rather than architectural advantages. For more details, please refer to \cref{sec:appendix}.

\paragraph{Evaluation} We employ standard evaluation metrics \citep{chen2019crowd,liu2023intentionawarerobotcrowd} to assess the robot's performance in navigation tasks. These metrics include \textit{success rate} (SR), which measures the proportion of episodes where the robot reaches its goal within a set time limit; \textit{collision rate} (CR), which calculates the fraction of episodes where the robot collides with any human; \textit{timeout rate} (TR), representing the ratio of episodes where the robot fails to reach the goal within the time limit; \textit{navigation time} (NT), which averages the time taken by the robot to reach the goal over successful episodes; \textit{path length} (PL), measuring the total distance traveled by the robot across all episodes, including collisions and timeouts; \textit{intrusion time ratio} (ITR), which tracks the fraction of time the robot intrudes into the predicted future positions of humans, triggering danger events; and \textit{social distance} (SD), which captures the average minimal distance between the robot and any human during the episode. These metrics provide a comprehensive view of the robot's ability to navigate efficiently, safely, and socially within dynamic environments. More detailed explanations can be found in the \cref{subsec:eval_metrics}.

\subsection{Results}
\label{sec:results}

The results presented in \cref{tab:main} demonstrate the effectiveness of \our{} in robot navigation, particularly in the context of automated reward function design. By comparing the performance of CrowdNav++ with that of \our{}, we show that a significant improvement in performance can be achieved by changing only the reward function, while leaving the underlying model architecture unchanged. This highlights the critical role that reward design plays in the success of reinforcement learning agents, validating the effectiveness of our automated reward function design process.

Additionally, when compared to other state-of-the-art reward function design methods, \our{} consistently outperforms both Eureka and CARD in terms of SR, CR, and TR. Specifically, \our{} achieves the highest success rate and the lowest collision and timeout rates, demonstrating that our approach not only enhances the model’s performance but also offers a more efficient and effective reward design process. These results further reinforce the efficacy of \our{} as a robust solution for automated reward function design in robot navigation tasks.

\subsection{Analysis}
\label{sec:analysis}

\begin{figure*}[t!]
    \centering
    \includegraphics[width=\linewidth]{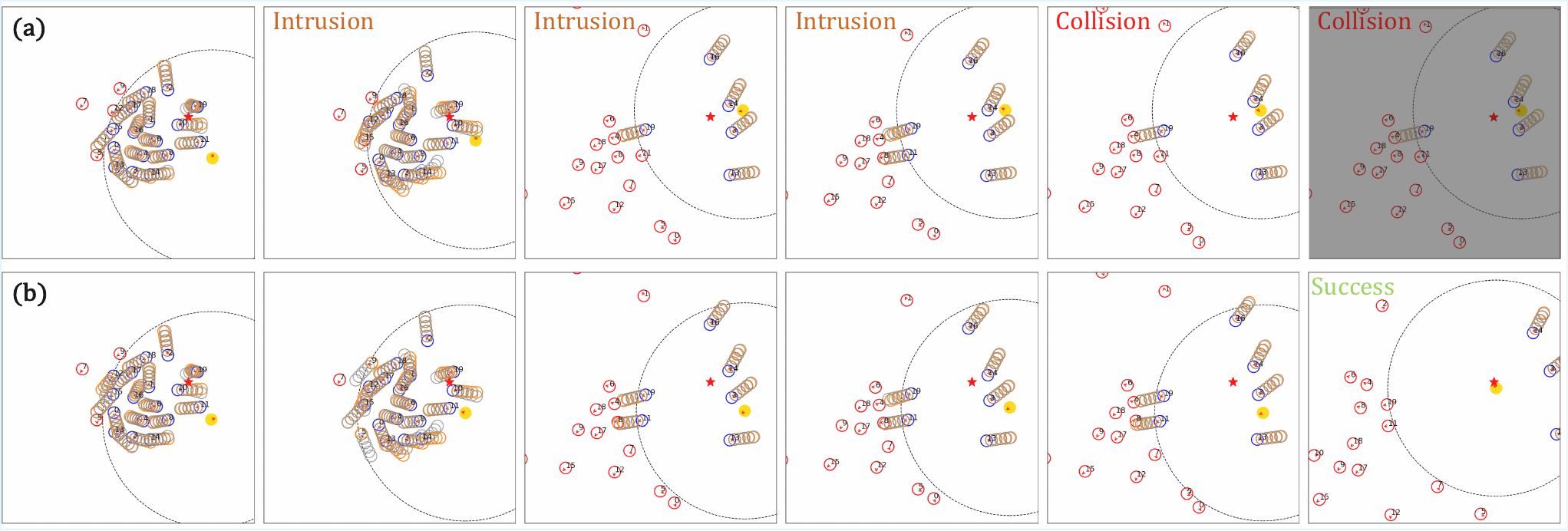}

    \caption{Navigation behavior comparison in dense crowd scenarios. Row (a) shows baseline policy (CrowdNav++) attempting aggressive navigation through crowds, resulting in intrusion events and collisions. Row (b) demonstrates \our{}'s evolved reward function achieving successful navigation through intelligent collision avoidance and efficient path planning.}
    \vspace{-4mm}
    \label{fig:visualization}
\end{figure*}

After evaluating the overall effectiveness of \our{}, this section addresses three key questions that further highlight the core contributions of our work: (1) What is the impact of each stage in the \our{} framework? (2) Why does using the proxy model to design the reward function work, and why is it a reasonable approach? (3) How efficient is \our{} in terms of the speed at which it can design reward functions?

\input{tables/ablation}

\paragraph{Ablation Study.}
The ablation study in \cref{tab:ablation} provides insights into the impact of each stage in the \our{} framework. As expected, the sequential nature of our design leads to cumulative improvements in performance, with each stage contributing progressively to the effectiveness of the reward function. Notably, Stage I already delivers significant improvement over the baseline method, CrowdNav++. Stage I+II further enhances the reward function, yielding a notable reduction in both CR and TR, while simultaneously improving the SR. The final \our{} model, which integrates all three stages, achieves the best results in all metrics, underscoring the value of the full framework.

An important observation from this study is that even after Stage II, the reward function designed by \our{} surpasses the performance of CARD, a SOTA reward design method. This result implicitly highlights the power and effectiveness of our proxy model in Stage I. Despite relying on simple proxies and without full policy training, the proxy-guided reward design already outperforms methods like CARD rely on more complex, iterative refinements. This emphasizes that even coarse initial reward functions, derived from the proxy model, can be highly effective when properly structured and fine-tuned by subsequent stages in the framework. This finding not only reinforces the efficiency of \our{} but also validates the critical role proxy models play in facilitating the reward design process, making it both fast and effective.

\begin{wrapfigure}{r}{0.45\textwidth}
    \centering
    \includegraphics[width=0.42\textwidth]{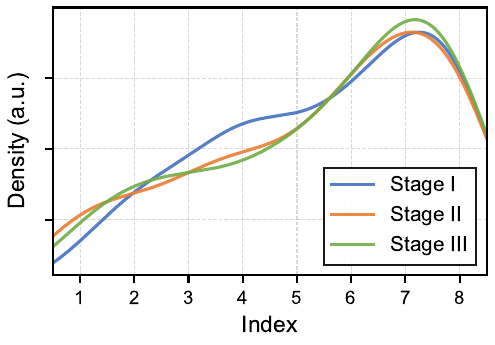}
    \caption{Performance distribution consistency across three progressive stages. All stages concentrate candidates in the high-performance region, with Stage II and Stage III showing nearly identical distributions, validating that lightweight proxy training predicts full-scale training outcomes.}
    \label{fig:proxy-consitency-validation}
    \vspace{-8mm}
\end{wrapfigure}

\paragraph{Proxy Consistency Validation.}
A key assumption underlying \our{}'s efficiency is that reward function rankings from early-stage proxies correlate with later full-training evaluations. To test this, we visualize performance distributions across stages (\cref{fig:proxy-consitency-validation}). All stages exhibit similar density patterns concentrated in the high-performance region, with Stage~II and Stage~III showing nearly identical distributions. This confirms that lightweight proxy training captures characteristics predictive of full-scale outcomes \citep{kaplan2020scaling,peherstorfer2018survey}, validating our progressive evaluation strategy.
\vspace{-2mm}

\paragraph{Computational Efficiency Analysis.}
We analyze the computational cost of \our{} to highlight the efficiency gains from its progressive three-stage framework. \our{} amortizes evaluation cost by triaging candidates through increasingly faithful signals: Stage~I performs analytic screening with no policy training, Stage~II leverages lightweight proxy policies for rapid iteration, and only a small subset proceeds to Stage~III for full-policy training. This organization concentrates expensive computation on promising candidates, is naturally parallelizable across candidates, and allows larger populations or more seeds under a fixed budget.

In contrast to approaches that train a full model for every candidate (e.g., CARD and Eureka), \our{} commits to full training only at the final stage, yielding substantially fewer costly runs while maintaining accuracy. Consistent with this design, ablations show that performance after Stage~II is already competitive with strong baselines, indicating that most search can be resolved before full training.
\vspace{-2mm}

\paragraph{Case Study and Qualitative Analysis.}
Through examining the evolved reward functions, we identify several distinctive design patterns. First, evolved rewards implement \textit{distance-dependent safety penalties} where the robot receives stronger penalties when very close to humans and gentler penalties when farther away, allowing aggressive navigation in open space while maintaining cautious behavior near obstacles. Second, they incorporate \textit{proximity-aware goal prioritization} where the importance of moving directly toward the goal increases as the robot approaches its destination, encouraging flexible detours when far away while demanding direct motion in the final approach. Third, evolved functions include \textit{path deviation penalties} that discourage unnecessary wandering and encourage the robot to stay near the geometrically optimal trajectory.

\cref{fig:visualization} illustrates how these patterns translate to superior navigation. Row (a) shows a baseline policy aggressively navigating through dense crowds, resulting in intrusion events and collisions. Row (b) demonstrates our evolved reward in an identical scenario, where the robot exhibits: (1) early crowd detection triggering wider berths around human clusters; (2) flexible detours enabled by proximity-aware prioritization while maintaining safe margins; and (3) smooth, predictable trajectories from path deviation penalties. These components work synergistically, resulting in successful navigation.
\vspace{-1mm}

%% file: tables/ablation.tex
\begin{table}[ht]
\centering
\vspace{-2mm}
\caption{Ablation study on \our{}'s three-stage framework. Each stage progressively improves performance, demonstrating cumulative benefits of the warm-up-boost pipeline.}
\vspace{2mm}
\label{tab:ablation}
\small
\begin{tabular}{l|ccc}
\toprule
\textbf{Method} & \textbf{SR}$\uparrow$ & \textbf{CR}$\downarrow$ & \textbf{TR}$\downarrow$ \\
\midrule
CrowdNav++    & $89.35\std{0.68}$ & $7.88\std{0.54}$ & $2.77\std{0.13}$  \\
\midrule
EvoNav (Stage I)  & $90.11\std{0.60}$ & $7.60\std{0.48}$ & $2.60\std{0.11}$ \\
EvoNav (Stage I+II) & $92.93\std{0.58}$ & $5.85\std{0.42}$ & $\mathbf{1.35}\std{0.10}$ \\
EvoNav (Full) & $\mathbf{93.22}\std{0.61}$ & $\mathbf{4.47}\std{0.42}$ & ${2.31}\std{0.10}$ \\
\bottomrule
\end{tabular}
\end{table}

%% file: sections/06-conclusion.tex
\section{Conclusion}
\label{sec:conclusion}
\vspace{-1mm}
In this paper, we presented \our{}, an evolutionary framework that automates reward function design for robot navigation using large language models. By introducing a progressive three-stage approach, we addressed the challenge of computationally expensive policy evaluations, efficiently navigating the reward space with inexpensive proxies in the early stages and refining the designs through full-scale training. Our experiments demonstrate that \our{} produces competitive navigation policies with fewer training iterations compared to traditional evolutionary methods. We also validate the proxy consistency hypothesis, showing that early-stage rankings predict later-stage performance, thus confirming the effectiveness of our approach. Overall, \our{} offers an efficient solution for automated reward design in robot navigation, with potential applicability to other reinforcement learning tasks and broader domains requiring scalable, cost-effective solutions.

%% file: appendix/a-algorithm.tex
\section{Algorithm}
\label{sec:algorithm}

\begin{algorithm}[h]
\caption{\our{} Framework}
\label{alg:evonav}
\begin{algorithmic}[1]
\STATE \textbf{Input:} LLM, seed function, dataset $\mathcal{D}$, population size $N$, Stage I generations $G_1$, Stage II rounds $G_2$, Stage III rounds $G_3$, training steps $(K_2, K_3)$ (see \cref{tab:framework_params} for values)

\STATE \textbf{Initialization:} $\mathcal{P}^{(0)} \gets$ LLM generates $N$ candidates from seed

\STATE \textbf{// Stage I: Evolution via Analytical Rules}
\FOR{$g = 1$ to $G_1$}
    \FOR{$r \in \mathcal{P}^{(g-1)}$}
        \STATE Compute $\text{Score}_1(r)$ on $\mathcal{D}$
    \ENDFOR
    \STATE Produce ranking $\mathcal{R}_1^{(g)}$ by sorting on $\text{Score}_1$
    \STATE $\mathcal{P}^{(g)} \gets$ LLM evolution (mutation/crossover/random) based on $\{\text{Score}_1(r)\}$
\ENDFOR
\STATE $\mathcal{P}^{(G_2)} \gets \mathcal{P}^{(G_1)}$

\STATE \textbf{// Stage II: Lightweight Proxy Model Rollouts}
\FOR{$g = 1$ to $G_2$}
    \FOR{$r \in \mathcal{P}^{(G_2)}$}
        \STATE Train lightweight policy $\pi_\phi^{\text{proxy}}$ with $r$ for $K_2$ steps
        \STATE Measure metrics $M(r) = (SR, NT, PL, ITR, SD)$
        \STATE $\mathcal{P}^{(g)} \gets$  LLM refines $r$ based on $M(r)$
    \ENDFOR
\ENDFOR
\STATE Produce ranking $\mathcal{R}_2$ via LLM evaluation of $\{M(r)\}$
\STATE $\mathcal{P}^{(G_3)} \gets \mathcal{P}^{(G_2)}$

\STATE \textbf{// Stage III: Final Refinement}
\FOR{$g = 1$ to $G_3$}
    \FOR{$r \in \mathcal{P}^{(G_3)}$}
        \STATE Train full policy with $r$ for $K_3$ steps
        \STATE Measure metrics $M(r) = (SR, NT, PL, ITR, SD)$
        \STATE $\mathcal{P}^{(g)} \gets$  LLM refines $r$ based on $M(r)$
    \ENDFOR
\ENDFOR
\STATE Produce final ranking $\mathcal{R}_3$ via LLM evaluation of $\{M(r)\}$

\STATE \textbf{Return:} Final candidates $\mathcal{P}^{(G_3)}$ and ranking $\mathcal{R}_3$
\end{algorithmic}
\end{algorithm}

%% file: appendix/b-hyperparameters.tex
\section{MDP Formulation}
\label{sec:mdp}

We model the navigation problem as a finite-horizon Markov Decision Process (MDP) $\mathcal{M}=(\mathcal{S},\mathcal{A},P,R_{\theta},\gamma,\rho_{0})$ over time $t\in\{0,\ldots,T-1\}$.

The transition kernel $P(s_{t+1}\mid s_t,a_t)$ is induced by the robot kinematics and the human motion model of the environment. The per-step reward is $R_{\theta}(s_t,a_t,s_{t+1})$, where $\theta$ parameterizes the reward design that will be optimized by \our{}. A (stochastic) policy $\pi_{\phi}(a\mid s)$ selects actions given states, and the objective is to maximize the expected discounted return
\[
J(\pi_{\phi};R_{\theta})=\mathbb{E}_{\substack{s_0\sim\rho_0,\;a_t\sim\pi_{\phi}(\cdot\mid s_t),\\ s_{t+1}\sim P(\cdot\mid s_t,a_t)}}\Big[\sum_{t=0}^{T-1}\gamma^{t}\,R_{\theta}(s_t,a_t,s_{t+1})\Big],
\]
where $\gamma\in[0,1)$ and $\rho_0$ is the initial-state distribution.

\section{Hyperparameters and Implementation Details}
\label{sec:appendix}


\begin{table}[h!]
\centering
\caption{Framework Architecture Parameters}
\label{tab:framework_params}
\begin{tabular}{lcc}
\toprule
\textbf{Parameter} & \textbf{Symbol} & \textbf{Value} \\
\midrule
Population Size & $N$ & 8 \\
Stage I Generations & $G_1$ & 10 \\
Stage II Refinement Rounds & $G_2$ & 16 \\
Stage III Refinement Rounds & $G_3$ & 3 \\
\bottomrule
\end{tabular}
\end{table}

\subsection{Stage I: Evolution via Analytical Rules}

\cref{tab:stage1_params} lists the parameters for the analytical rule-based evaluation stage.

\begin{table}[h!]
\centering
\caption{Stage I Parameters}
\label{tab:stage1_params}
\begin{tabular}{lcc}
\toprule
\textbf{Parameter} & \textbf{Symbol} & \textbf{Value} \\
\midrule
Number of Scenarios & $M$ & 100 \\
Trajectories per Scenario & $N_{\text{traj}}$ & 10 \\
Frames per Scenario & $F_j$ & $\approx 200$ \\
\bottomrule
\end{tabular}
\end{table}

\subsection{Stage II: Lightweight Proxy Model Rollouts}

\cref{tab:stage2_params} shows the training and evaluation parameters for the lightweight proxy model stage.

\begin{table}[h!]
\centering
\caption{Stage II Parameters}
\label{tab:stage2_params}
\begin{tabular}{lcc}
\toprule
\textbf{Parameter} & \textbf{Symbol} & \textbf{Value} \\
\midrule
Training Steps & $K_2$ & 8,000 \\
Evaluation Episodes & $E_2$ & 50 \\
Episode Horizon & $T_{\text{short}}$ & 100 steps \\
\bottomrule
\end{tabular}
\end{table}

\subsection{Stage III: Final Refinement}

\cref{tab:stage3_params} presents the parameters for full-scale policy training and comprehensive evaluation.

\begin{table}[h!]
\centering
\caption{Stage III Parameters}
\label{tab:stage3_params}
\begin{tabular}{lcc}
\toprule
\textbf{Parameter} & \textbf{Symbol} & \textbf{Value} \\
\midrule
Training Steps & $K_3$ & $10^7$ \\
Evaluation Episodes & $E_3$ & 500 \\
Human Counts Tested & $H$ & $\{5, 10, 15, 20\}$ \\
\bottomrule
\end{tabular}
\end{table}

\subsection{MDP and Environment Parameters}

\cref{tab:mdp_params} summarizes the core Markov Decision Process parameters and environment settings used in robot navigation.

\begin{table}[h!]
\centering
\caption{MDP and Environment Parameters}
\label{tab:mdp_params}
\begin{tabular}{lcc}
\toprule
\textbf{Parameter} & \textbf{Symbol} & \textbf{Description} \\
\midrule
Episode Horizon & $T$ & Maximum time steps \\
Discount Factor & $\gamma$ & $\gamma \in [0,1)$ \\
Maximum Velocity & $v_{\max}$ & Maximum action norm \\
Robot Radius & $r_0$ & Disc footprint radius \\
Goal Threshold & $\varepsilon_{\text{goal}}$ & Success distance \\
Time Step & $\Delta t$ & Discretization interval \\
Human Radius & $r_h$ & Human footprint radius \\
\bottomrule
\end{tabular}
\end{table}

\subsection{Evaluation Metrics}
\label{subsec:eval_metrics}

The performance of each candidate reward function is evaluated using seven key metrics. Success Rate (SR) measures the percentage of episodes in which the robot successfully reaches the goal within the time limit. Collision Rate (CR) quantifies the percentage of episodes involving collisions with humans or obstacles. Timeout Rate (TR) represents the percentage of episodes where the robot fails to reach the goal before the time limit expires. Navigation Time (NT) denotes the average time required for the robot to reach the goal in successful episodes. Path Length (PL) reflects the average total distance traveled by the robot across all episodes, including those ending in collisions or timeouts. Intrusion Time Ratio (ITR) measures the proportion of time the robot spends intruding into the predicted personal spaces of nearby humans. Social Distance (SD) captures the average minimum distance maintained between the robot and humans during navigation, reflecting the level of social compliance and comfort.

\subsection{LLM Configuration}

The large language model used for reward function generation and refinement employs structured prompting with: Evolutionary Operations: Mutation (modifying existing components), crossover (combining parent functions), and random initialization (maintaining diversity); Feedback Format: Stage I uses correlation scores; Stages II and III use multi-objective metric tuples $M(r) = (SR, NT, PL, ITR, SD)$; Code Interface: Standardized \texttt{compute\_reward(inst, traj)} signature for all generated reward functions.

\subsection{LLM Ablation Study}

We evaluate different LLMs while keeping all other framework components identical. \cref{tab:llm_comparison} shows the success rate for each model.

\begin{table}[h!]
\centering
\caption{Success Rate Comparison Across Different LLMs}
\label{tab:llm_comparison}
\begin{tabular}{lcc}
\toprule
\textbf{LLM} & \textbf{Size} & \textbf{SR (\%)} \\
\midrule
Llama-3.1-70B & 70B & 88.5 \\
GPT-4o-mini & Closed & 86.2 \\
Llama-3.1-405B & 405B & 90.1 \\
GPT-4o & Closed & 90.8 \\
GPT-OSS-120B & 120B & \textbf{91.2} \\
\midrule
CrowdNav++ (Baseline) & - & 84.3 \\
\bottomrule
\end{tabular}
\vspace{2mm}

\end{table}

\begin{table}[h!]
\centering
\caption{Cost Comparison Between Methods}
\label{tab:cost_simple}
\begin{tabular}{lccc}
\toprule
\textbf{Method} & \textbf{GPU Cost} & \textbf{LLM Cost} & \textbf{Total} \\
\midrule
CARD & 36 hrs (\$18.00) & 72K tokens (\$0.02) & \$18.02 \\
\our{} & 9.2 hrs (\$4.60) & 720K tokens (\$0.21) & \$4.81 \\
\midrule
\textbf{Savings} & \textbf{74\%} & - & \textbf{73\%} \\
\bottomrule
\end{tabular}
\vspace{2mm}

\end{table}

Key findings: (1) Performance varies with LLM capability, ranging from 86.2\% to 91.2\%; (2) All LLMs outperform the manually designed baseline (84.3\%), validating automated reward design; (3) The three-stage framework consistently benefits all models, demonstrating robustness across different LLM choices.

\subsection{Proxy Consistency Hypothesis Parameters}

For validating the proxy consistency hypothesis: Correlation Threshold: $\delta > 0$ for meaningful cross-stage alignment; Measured Correlations: Spearman rank correlation $\rho_{ij}$ between stages $i$ and $j$; Expected Performance: $\rho_{12}, \rho_{23} > 0.5$ indicates effective proxy guidance.

%% file: appendix/c-prompts.tex
\section{Prompts}
\subsection{Initial Population}
The following prompt is used for generating the first set of diverse reward function candidates, which combines a system-level instruction and a user-level task definition.
\begin{promptbox}{System Prompt}
\begin{Verbatim}[fontsize=\small]
You are an expert in reinforcement learning and robot navigation. Your goal is to design reward functions that effectively guide a robot toward safe, efficient, and socially compliant navigation behaviors.

Return **only** valid Python code enclosed within a fenced code block. The code must be fully executable and should not include comments or explanations outside the block.
\end{Verbatim}
\end{promptbox}

\begin{promptbox}{User Prompt}
\begin{Verbatim}[fontsize=\small]
Please write a Python function named {func_name} for a robot navigation task in a crowded environment.

Task Description:
- The function's goal is to output a scalar reward value based on the robot's current state, guiding it to its goal while avoiding collisions with dynamic human agents.

Function Interface:
- Inputs:
  - inst: A scenario object containing static information like robot_start, robot_goal, and human_trajectories.
  - traj: The robot's trajectory history, represented as a list of positions. The last element is the current state.
- Output:
  - A single scalar (float) representing the reward for the current state or action.

Design Principles:
- Goal-Progress: The function should reward progress toward the robot_goal.
- Collision Avoidance: The function must penalize states that are too close to humans or result in a collision.
- Interpretability: The code should be well-commented, and variable names should be clear to facilitate human understanding.

Constraints (CRITICAL):
- Hyperparameters: All tuning parameters (e.g., weights, constants) must be defined as local variables inside the function body. Do not add them as function arguments.
- Output Format: Your response must contain only the Python function within a single code block. Do not include any explanatory text or print statements.
\end{Verbatim}
\end{promptbox}

\subsection{Evolutionary Operations}
The following prompt is used for evolving the population of reward functions by guiding the LLM to create new functions by mutation or crossover existing ones based on performance feedback.
\begin{promptbox}{Crossover}
\begin{Verbatim}[fontsize=\small]
You are a reward function architect. Your task is to synthesize a new function by combining the complementary strengths of two parent functions. Below are two candidate reward functions and a textual reflection. Use them to synthesize an improved hybrid design that combines the strengths of both functions while addressing weaknesses noted in the reflection.

Parent A:
- {code_A}

Parent B:
- {code_B}

Reflection:
- {reflection}.

Synthesis Task:
- Based on the provided reflection, write a new, improved function `{func_name}_v2` that merges the superior safety features from Parent A with the efficiency-promoting logic from Parent B.
\end{Verbatim}
\end{promptbox}

\begin{promptbox}{Mutation}
\begin{Verbatim}[fontsize=\small]
You are a reward function optimizer. Your task is to perform a targeted mutation on a high-performing function to address a specific weakness identified in the reflection. A prior reflection summarizing performance feedback is provided below. Use it to locally improve the current elite reward function by small, targeted edits that preserve its strengths while addressing weaknesses.

Prior Reflection:
- {reflection}

High-Performing Code to Mutate:
- {func_signature}
- {elitist_code}

Mutation Task:
- Based on the reflection, create a mutated function `{func_name}_v2`. Make a minimal, precise change to the code to address the identified weakness without degrading its existing strengths.
\end{Verbatim}
\end{promptbox}

\subsection{Reward Function Refinement}

\begin{promptbox}{System Prompt}
\begin{Verbatim}[fontsize=\small]
You are a senior researcher in robot motion planning and reinforcement learning. Rewrite the provided reward function to produce a smooth, dense, and numerically stable per-frame signal that differentiates between successful, safe, and inefficient navigation behaviors.

Requirements:
- Keep the original function signature.
- Maintain O(T) computational complexity.
- Ensure bounded reward magnitudes and graceful handling of incomplete trajectories.
- Output only valid Python code (no markdown fences or commentary).
\end{Verbatim}
\end{promptbox}

\begin{promptbox}{User Prompt}
\begin{Verbatim}[fontsize=\small]
Current score (best so far): {last_score:.4f} (higher is better, approx. range -1 to 1)

Core components to ensure (abstract, unordered):
- Progressive advancement signal
- Safety differentiation
- Terminal / completion handling
- Efficiency (avoid needless detours)
- Stability (avoid erratic frame jumps)

Guidelines:
Strengthen discrimination between clearly successful, safe, efficient progress and poor / unsafe / stagnant behavior. Maintain:
- Dense per-frame shaping (not a single terminal spike)
- Bounded, numerically stable magnitudes (avoid runaway growth)
- Graceful handling of incomplete trajectories
Discourage unsafe or aimless motion and excessive oscillation without over-penalizing reasonable detours. Keep the logic straightforward and linear-time.

Keep all existing effective terms (progress-to-goal, safety distance shaping, smoothness, efficiency) unless there is a concrete numerical reason to adjust them.
Make only minimal, localized changes that strictly improve discriminative sharpness without removing previously working logic.
Avoid producing nearly constant rewards across different frames or trajectories; variance should reflect qualitative behavioral differences.

Focus note: {feedback}

{extra_context_if_any}

Revise the function below.
Maintain the original signature and return a meaningful per-frame shaping signal aggregated appropriately.
Return **only** the updated function definition (no additional text).

{current_code}
\end{Verbatim}
\end{promptbox}

\subsection{External Knowledge}

\begin{promptbox}{External Knowledge}
\begin{Verbatim}[fontsize=\small]
# External Knowledge for Reward Function Design

## Task Definition

- Domain: Crowd-robot navigation in continuous 2D environments
- Dataset: Synthetic environments derived from popular benchmarks, where human agents follow realistic social trajectories
- Input:
  - Current robot position $(x_t, y_t)$
  - Goal position $g = (x_g, y_g)$
  - Human positions $\{p_h(t)\}_{h=1}^H$ within the robot’s observation field
- Output:
  - A **reward function** $r$ that maps each navigation state to a scalar reward signal used to train reinforcement learning policies (e.g., PPO)
- Objective:
  - Encourage goal-directed progress
  - Penalize collisions and unsafe proximity
  - Promote smooth, efficient, and socially compliant motion

## Evaluation Metrics
- SR (Success Rate): Percentage of episodes where the robot successfully reaches the goal within the time limit
- CR (Collision Rate): Percentage of episodes involving collisions with humans or obstacles
- TR (Timeout Rate): Percentage of episodes where the robot fails to reach the goal within the time limit
- NT (Navigation Time): Average time taken to reach the goal in successful episodes
- PL (Path Length): Average total distance traveled, including during collisions and timeouts
- ITR (Intrusion Time Ratio): Fraction of time the robot intrudes into humans' predicted positions, triggering danger events
- SD (Social Distance): Average minimum distance between the robot and nearby humans during navigation
\end{Verbatim}
\end{promptbox}

\subsection{Seed Function}

\begin{lstlisting}[caption={}, label={lst: interactions}, language=Python, style=heuristicstyle]]
import numpy as np
def seed_reward_func(inst, traj):
    """
    Args:
        - inst: single instance, with the shape of
        - traj: single trajectory
    """
    start_x, start_y = inst["robot_start"]
    goal_x, goal_y = inst["robot_goal"]
    curr_x, curr_y = traj[-1]
    prev_x, prev_y = traj[-2]
    robot_radius = traj[-1][-1]
    reach_goal_flag = np.linalg.norm([curr_x - goal_x, curr_y - goal_y]) < robot_radius
    collision_flag = False
    for human_traj in inst["human_trajectories"]:
        for human_x, human_y in human_traj:
            if np.linalg.norm([curr_x - human_x, curr_y - human_y]) < robot_radius:
                collision_flag = True
                break
    if reach_goal_flag:
        return 10
    elif collision_flag:
        return -20
    else:
        curr_dist = np.linalg.norm([curr_x - prev_x, curr_y - prev_y])
        prev_dist = np.linalg.norm([prev_x - start_x, prev_y - start_y])
        return 2 * (prev_dist - curr_dist)
\end{lstlisting}